\def\year{2022}\relax
\documentclass[letterpaper]{article}
\usepackage{aaai}
\usepackage{times}
\usepackage{helvet}
\usepackage{courier}
\usepackage[hyphens]{url} 
\usepackage{microtype}
\usepackage{graphicx}
\usepackage{subfigure}
\usepackage{xspace}
\usepackage{lineno}
\usepackage{booktabs}
\newcommand{\renyi}{R\'enyi\xspace}
\newcommand{\nosection}[1]{\vspace{2pt}\noindent\textbf{#1.}}
\newtheorem{definition}{Definition}
\usepackage{multirow}
\usepackage{wrapfig,lipsum,booktabs}
\newtheorem{theorem}{Theorem}
\newtheorem{lemma}{Lemma}
\usepackage{amsfonts}       % blackboard math symbols
\usepackage{amssymb}
\usepackage{amsmath}
\usepackage{soul}
\usepackage{xcolor}
\usepackage{wrapfig,lipsum,booktabs}
\begin{document}
% \linenumbers
\title{Generalization Bounds for Stochastic Gradient Langevin Dynamics: \\
A Unified View via Information Leakage Analysis}
\author{Bingzhe Wu$^1$  Zhicong Liang$^2$ Yatao Bian $^1$ ChaoChao Chen $^3$ Junzhou Huang $^1$ Yuan Yao $^2$ \\ $^1$Tencent AI Lab $^2$HongKong University of Science and Technology $^3$Zhejiang University\\
}
\maketitle
\begin{abstract}
% The generalization bounds of the non-convex empirical risk minimization paradigm using Stochastic Gradient Langevin Dynamics (SGLD) have been extensively studied recently. Several theoretical frameworks were presented to study this problem from different perspectives, such as information theory and stability. In this paper, we present a unified view from privacy leakage analysis to study the generalization bounds of SGLD. An associated theoretical framework is also presented for re-deriving previous theoretical results under the viewpoint of privacy leakage. 
% Our framework can cover a wide range of previous theoretical works that focus on different aspects. Besides theoretical findings, we also perform various numerical studies to empirically evaluate the information leakage of SGLD. Furthermore, our theoretical and empirical results could  also provide explanations for the prior works that study the membership privacy of SGLD.

Recently, generalization bounds of the non-convex empirical risk minimization paradigm using Stochastic Gradient Langevin Dynamics (SGLD) have been extensively studied. Several theoretical frameworks have been presented to study this problem from different perspectives, such as information theory and stability. In this paper, we present a unified view from privacy leakage analysis to investigate the generalization bounds of SGLD, along with a theoretical framework  for re-deriving previous  results in a  succinct manner. 
    Aside from theoretical findings, we  conduct various numerical studies to empirically assess the information leakage issue of SGLD. Additionally, our theoretical and empirical results provide explanations for  prior works that study the membership privacy of SGLD.
\end{abstract}

\section{Introduction}
One of the fundamental problems in deep learning is to characterize generalization bounds of non-convex stochastic optimization in the empirical risk minimization (ERM) setting~\cite{colt_sgld_gen,benjo_generalization,uniform_convergence,Arora0NZ18}. 
Some traditional approaches study the generalization bounds by measuring the complexity of the hypothesis space in a data-dependent manner (e.g., the VC dimension~\cite{vapnik1998adaptive} and the Rademacher complexity~\cite{rademacher}).
Nevertheless, directly adopting these complexity measures often fail to characterize the generalization ability of various learning algorithms for training deep neural networks~(DNNs), in which the algorithm explores the hypothesis space in an algorithm-dependent manner~\cite{ZhangBHRV17,colt_sgld_gen}. Prior works have shown various empirical evidence from the observation that the hyper-parameters of stochastic gradient methods could significantly affect the generalization ability of the learned DNN models~\cite{colt_sgld_gen,large_batch_training}. For example, the large batch size training
often fails to generalize to the test dataset well while the small batch size could introduce more stochastic noise as an implicit regularizer, which further improves the generalization ability of the learned DNN models~\cite{large_batch_training}.

The aforementioned issues are driving the research on algorithm-dependent generalization bounds in the deep learning context.
Specifically, the generalization bound of Stochastic Gradient Langevin Dynamics (SGLD) in the non-convex setting has drawn increasing attention from the research community. Previous works have demonstrated various generalization bounds of SGLD from different theoretical perspectives~\cite{colt_sgld_gen,isit_sgld,nips_sgld_info}. 
% The major theoretical tools employed in these works include \emph{stability-based theory} and \emph{information theory}.
\emph{Stability-based theory} and \emph{information theory}  are the two most important theoretical tools used in these works. 
A seminal work that leverages the stability-based theory to obtain the generalization bound of SGLD is conducted by Mou et al.~\cite{colt_sgld_gen}. 
The key idea of this work is to prove the uniform stability of SGLD concerning the squared Hellinger distance and further derive the bound via the connection between uniform stability and the expected generalization error~\cite{shwartz_stable,colt_sgld_gen}.
A follow-up work improves the generalization bound by introducing the Bayes stability notion~\cite{nips_sgld_info}. 
Another line of research works study this problem based on information theory. The core idea is to bound the generalization error via some information measures. For example, an early work \cite{XuR17} proposes to bound the expected generalization error in terms of the mutual information between the training dataset and the output hypothesis of the learning algorithm. Bu et al.~\cite{isit_sgld} introduce the individual sample mutual information and link it to the expected generalization error. 

Despite these theoretical results, the connections amongst these lines of research and how to build a unified understanding of them remain largely unexplored. 
In this paper, we present a unified view to study the generalization ability of SGLD via analyzing the privacy leakage of SGLD. Our work is inspired by
a number of recent works~\cite{p3sgd,gan_privacy,yuxiang_dp_stable}, which show that there are both theoretical and empirical connections between the privacy leakage and the generalization error. Intuitively, \emph{``improving the generalization ability''} and \emph{``reducing the individuals' privacy leakage''} share the same goal of encouraging a model to learn the population's features instead of memorizing the features of each individual~\cite{dwork_dp,p3sgd}. 
Theoretically, previous results have shown that the commonly used privacy notion of differential privacy~\cite{dwork_dp} may provide the property of uniform stability, which can be further related to the bound of the expected generalization error. 
Unfortunately, the notion of differential privacy is too strict to be directly adopted for obtaining reasonable generalization bounds for SGLD, i.e., the higher privacy budget of differential privacy would lead to vacuous bound while lower privacy budget will cause a low model performance. 

In this paper, we propose using a relaxed privacy notion based on 
\renyi divergence to analyze the privacy leakage of SGLD.
Specifically, under some mild assumptions (e.g., the boundness of the loss function), we first compute the bound of the \renyi divergence between weight parameters learned using two adjacent datasets at each iteration, i.e., the ``leave-one-out'' \renyi divergence. 
We could then use the chain rule of the \renyi divergence to compose these bounds of different iterations for deriving the total bound of the final weights learned using SGLD. 
Based on such a bound on the \renyi divergence, we have a simple and unified approach to obtain various theoretical bounds in previous studies. 
For example, the foundation of the work in~\cite{colt_sgld_gen} is uniform stability with respect to the Hellinger distance, which implies the boundedness of the Hellinger distance between the hypothesis (i.e., weight parameters) learned using two adjacent datasets. 
Under our framework, the boundedness of the Hellinger distance can be obtained by the bound of the \renyi divergence, 
% thus we can further obtain the generalization bound of SGLD. 
thus can be further used to obtain the generalization bound of SGLD.
Moreover, our framework could also give rise to previous works~\cite{XuR17,isit_sgld} with an information-theoretic view and has the potential to improve previous generalization bounds by introducing tighter privacy notions. See more details and results in Section~\ref{sec:theorem}.

Besides the theoretical contributions, we also conduct experiments on several datasets where data privacy is desired to empirically evaluate the information leakage of SGLD against membership attacks. Specifically, we design different attacks (e.g., membership attacks) to evaluate the information leakages of models learned using SGLD and SGD. 
We empirically observe that SGLD could not only improve the generalization ability of a model but also reducing its membership information leakage for the training dataset.

\section{Theoretical Framework}
\label{sec:theorem}
\subsection{Preliminaries}
We denote $\mathcal{D}$ as the unknown population distribution on the sample space $\mathcal{Z}$ and denote
$\mathcal{W}$ as the parameter space of the hypothesis (e.g., a neural network). 
Considering a loss function $l:\mathcal{Z}\times \mathcal{W}\rightarrow \mathbb{R}$,
the goal of a learning algorithm is to find the parameter $w$ which minimizes the population risk $L_{\mathcal{D}}(w) = \mathbb{E}_{z\sim\mathcal{D}}l(z;w)$.
Empirically, the population distribution is always intractable, thus we turn to the empirical
risk minimization~(ERM). 
%In the paradigm of ERM, 
In the ERM paradigm, a training dataset $S=\{z_i\}^{n}_{i=1}$ is given and each
$z_i$ is sampled from $\mathcal{D}$ independently, i.e., $z_i{\overset{\text{i.i.d.}}{\sim}} \mathcal{D}$. In particular, $z_i=(x_i, y_i)$ is a sample-label pair in the setting of supervised learning. 
The aim of a learning algorithm in the ERM setting is to find the parameter which minimizes the empirical risk
$L_{\mathcal{S}}(w) = {\tiny \dfrac{1}{n}}\sum_{i=1}^{n}l(z;w)$.

In the above ERM paradigm, given a learning algorithm $\mathcal{A}$, we are interested in its expected \emph{generalization error (gap)}, which is the
expected difference between the empirical and population risk,
\begin{equation}
    \text{gen}(\mathcal{A},\mathcal{D}) = \mathbb{E}_{w\sim\mathcal{A},S\sim\mathcal{D}^{n}}[L_{\mathcal{D}}(w)-L_{\mathcal{S}}(w)],
\label{eq:gen_bound}
\end{equation}
where the expectation is taken over the randomness from both the sampling process of the training dataset  and the learning algorithm (e.g., stochastic error or the injected noises in SGLD).

\subsection{Privacy Leakage Analysis}
In this paper, we study the bound of the expected generalization error from the view of privacy analysis.
The learning algorithm $\mathcal{A}$  studied in this paper is SGLD, which can be seen as a noisy version of SGD. We denote $w_t$ as the output parameter at the $t$-th iteration. Following the prior work \cite{nips_sgld_info}, we employ the SGLD updating rule as follows in this paper:
\begin{equation}
    w_{t+1} = w_t-\alpha_t\partial_{w} L(\mathcal{B}_t, w_t)+\eta_t, \eta_t\sim\mathcal{N}(0, 2\alpha_t \mathbf{I})
    \label{eq:sgld_update}
\end{equation}
\begin{equation}
     L(\mathcal{B}_t, w_t)= \dfrac{1}{|B_t|}\sum_{ z_i \in \mathcal{B}_t}l(z_i, w_t)\text{,}
\label{eq:sgld_loss}
\end{equation}
where $\alpha_t$ is the step size. $\mathcal{N}(0, 2\alpha_t\mathbf{I})$ denotes the Gaussian distribution. $L(\mathcal{B}_t, w_t)$ is the loss function computed on a mini-batch $\mathcal{B}_t$ that is randomly selected from the whole dataset $S$. Due to the injected noise and the randomness from the mini-batch, the output parameter of SGLD
at each iteration can be seen as a random variable with the density function $p(w_{t+1}|w_{t}, S)$. The goal of this part is to measure the information leakage of the output of the learning algorithm $\mathcal{A}$, i.e.,
$p(w_T|S)$, where $T$ denotes the number of iterations. 

In this paper, motivated by recent advances in \renyi differential privacy~\cite{renyi_dp}, we propose to quantitatively evaluate the information 
leakage by measuring the difference between the output distributions computed on two adjacent datasets.
To do this, we first introduce the concept of \renyi divergence which is used for measuring the difference between two probability distributions:
\begin{definition}
\textbf{(\renyi divergence).} For two probability distributions $p$ and $q$ (probability density functions), the \renyi divergence of order $\lambda$ is defined as:
\begin{equation*}
D_\lambda(p\|q)\triangleq\frac{1}{\lambda-1}\log \mathbb{E}_{x\sim q}\left( \frac{p(x)}{q(x)}\right)^\lambda.
\end{equation*}
\label{def:renyi}
\end{definition}
We then introduce the concept of adjacent datasets: two datasets $S$ and $S'$ are adjacent when one can be obtained by removing a single element from the other. Without loss of generality, we assume $S'=S / \{z_n\}$. We further denote $w_t$ and $w'_t$ as the output of SGLD in the $t$-th iteration, which are computed using $S$ and $S'$, respectively.
We are actually interested in the quantity of $D_{\lambda}(p(w_T|S)\|p(w'_T|S'))$. The bound of this quantity can also be seen as the privacy cost in the notion of \renyi differential privacy~\cite{renyi_dp}.

%In our paper, the randomized mechanism is set to be the SGLD algorithm.

At a high level, the bound of $D_{\lambda}(p(w_T|S)\|p(w'_T|S'))$ can be obtained following two steps: (1) deriving the \renyi divergence between $p(w_t|w_{<t})$~\footnote{$w_{<t}$ refers to the sequence $(w_0, w_1, \cdots, w_{t-1})$.} and $p(w'_t|w'_{<t})$ at the $t$-th
iteration; (2) composing these divergences by the composition theory for the adaptive mechanisms (i.e., learning algorithms). In the case of SGLD, we note that $w_t$ only depends
on $w_{t-1}$, thus the conditional distribution $p(w_t|w_{<t})$ can be reduced to $p(w_t|w_{t-1})$.

For step (1), following prior works on analyzing the privacy leakage of DP-SGD~\cite{dp_dl}, we treat $p(w_t|S)$ and $p(w'_t|S')$ as Mixture-Gaussian distribution in the setting of SGLD, then we can explicitly compute the divergence and
obtain the following bound:
\begin{lemma}\label{lemma:one_step_bound}\textbf{(Leave-one-out Stability of \renyi Divergence)}
Suppose $p(w_t|w_{t-1})$ and $p(w'_t|w'_{t-1})$ are the probability distributions computed using SGLD in Eq.~(\ref{eq:sgld_update}) (\ref{eq:sgld_loss}) given $w_{t-1}$ and $w'_{t-1}$, and Poisson subsampling with sampling ratio $\tau$ is applied, then the R\'enyi divergence between these two distributions follows (denoting $\beta_t=\dfrac{1}{|\mathcal{B}_t|} $.):
    \begin{equation*}
        \begin{aligned}
            D_\lambda (p(w_{t}|w_{t-1}\| p(w'_{t}|w_{t-1})) \leq \dfrac{\lambda \alpha_t L^2}{n^2}
        \end{aligned}
    \end{equation*}
    given that  $\sigma^2\geq 0.53$ and $\lambda-1 \leq \frac{2}{3}\sigma^2\log \big(1/\lambda\tau(1+\sigma^2)\big)$, where $\sigma^2 := \frac{2}{ \alpha_t \beta_t^2 L^2}$, where $L$ is the bound of the gradient norm $\|\partial_w\ell(z,w)\|$.
\end{lemma} 
\textbf{Proof:}
Suppose that  we have:
\begin{equation*}
    \begin{aligned}
        w'_{t+1} 
        &\sim \sum_{\mathcal{B}'_t} p(\mathcal{B}'_t) \cdot \mathcal{N}\Big(w'_{t}-\alpha_{t} \partial_w L(\mathcal{B}'_{t}, w_{t}), 2\alpha_t \bf{I} \Big)
    \end{aligned}
\end{equation*}
where $\mathcal{B}'_{t}$ is the set of selected samples from $\mathcal{S}'$ with sampling ratio $\tau$.

Since $\mathcal{S} = \mathcal{S'}\cup \{z_n\}$, we have
\begin{equation*}
    \begin{aligned}
        w_{t+1}
        &\sim \quad \sum_{\mathcal{B}'_t} p(\mathcal{B}'_t) \Bigg( (1-\tau) \mathcal{N}\Big( w_{t} - \alpha_t \partial_w L(\mathcal{B}'_{t}, w_{t}), 2\alpha_t \textbf{I} \Big) \\&+  \tau \mathcal{N}\Big( w_{t} - \alpha_t \partial_w L(\mathcal{B}'_{t} \cup \{z_n\}, w_{t}), 2\alpha_t \textbf{I} \Big) \Bigg)
    \end{aligned}
\end{equation*}
For all order $\lambda$, R\'enyi divergence is quasi-convex \cite{van2014renyi},
\begin{equation*}
    \begin{aligned}
        &D_\lambda (p(w_{t+1}|w_t\| p(w'_{t+1}|w_t)) \leq \\ &\sup_{\mathcal{B}'_t} D_\lambda 
        \Bigg( (1-\tau) \mathcal{N} \Big(w_{t} - \alpha_t \partial_w \big( -\log p(w_{t}) \\
        &+ \beta_t \sum_{z_t \in \mathcal{B}'_t} \ell (z_i, w_{t} ) \big), 2\alpha_t \textbf{I} \Big)  \\
        & + \tau \mathcal{N}\Big( w_{t} - \alpha_t \partial_w \big( -\log p(w_{t}) \\
        &+ \beta_t \sum_{z_i \in \mathcal{B}_t'\cup \{z_n\}} \ell (z_i, w_{t}) \big), 2\alpha_t \textbf{I} \Big)  \Big\| \\ &\mathcal{N}\Big(w_{t}-\alpha_t \partial_w \big(-\log p(w_{t}) + \beta_t \sum_{z_t \in \mathcal{B}'_{t}} \ell (z_i, w_{t}) \big), 2\alpha_t \textbf{I} \Big) \Bigg)\\
    \end{aligned}
\end{equation*}
By the translation invariance property of R\'enyi divergence,
\begin{equation*}
    \begin{aligned}
        &D_\lambda (p(w_{t+1}|w_t\| p(w'_{t+1}|w_t)) 
        \leq\\
        &=  D_\lambda 
        \Bigg( (1-\tau) \mathcal{N} \Big(0 , 2\alpha_t \textbf{I} \Big) + \tau \mathcal{N}\Big( -\alpha_t \beta_t \partial_w \ell (z_n, w_{t}) \big), 2\alpha_t \textbf{I} \Big)  \Big\|\\ &\mathcal{N}\Big(0, 2\alpha_t \textbf{I} \Big)  \Bigg)\\
    \end{aligned}
\end{equation*}
where the last equality comes from the fact that the term on the right-hand side is independent of $\mathcal{B}'_t$.
Since the covariances are symmetric, we can assume that $-\alpha_t \beta_t \partial_w \ell (z_n, w_{t})=c \bf{e}_1$ for some constant $c\leq L$. The two distributions at hand are then both product distributions that are identical in all coordinates except the first. By additivity of R\'enyi divergence for product distributions \cite{van2014renyi}, we have 
\begin{equation*}
    \begin{aligned}
        &D_\lambda (p(w_{t+1}|w_t\| p(w'_{t+1}|w_t)) \leq \\& \sup_{c\leq C} D_\lambda 
        \Bigg( (1-\tau) \mathcal{N} \Big(0 , 2\alpha_t  \Big) + \tau \mathcal{N}\Big(\alpha_t \beta_t c, 2\alpha_t \Big)  \Big\|\mathcal{N}\Big(0, 2\alpha_t \Big)  \Bigg)\\
        &\leq D_\lambda 
        \Bigg( (1-\tau) \mathcal{N} \Big(0 , 2\alpha_t  \Big) + \tau \mathcal{N}\Big(\alpha_t \beta_t C, 2\alpha_t \Big)  \Big\|\mathcal{N}\Big(0, 2\alpha_t \Big)  \Bigg) \\
        &= D_\lambda 
        \Bigg( (1-\tau) \mathcal{N} \Big(0 , \sigma^2  \Big) + \tau \mathcal{N}\Big(1, \sigma^2 \Big)  \Big\|\mathcal{N}\Big(0, \sigma^2 \Big)  \Bigg) \\
    \end{aligned}
\end{equation*}
where we denote $\sigma^2 := \frac{2}{ \alpha_t \beta_t^2 L^2}$. Now we reduce our problem to the bounding of  R\'enyi divergence of Gaussian mechanism with Poisson subsampling:
\begin{equation*}
    \begin{aligned}
        &D_\lambda 
        \Bigg( (1-\tau) \mathcal{N} \Big(0 , \sigma^2  \Big) + \tau \mathcal{N}\Big(1, \sigma^2 \Big)  \Big\|\mathcal{N}\Big(0, \sigma^2 \Big)  \Bigg) \\
        &\leq \frac{1}{\lambda-1} \log \Bigg( (\lambda\tau-\tau+1)(1-\tau)^{\lambda-1} +\\& {\lambda \choose 2}(1-\tau)^{\lambda-2} \tau^2 e^{\rho(2)} + \sum_{j=3}^{\lambda} {\lambda \choose j}(1-\tau)^{\lambda-j} \tau^j e^{(j-1)\rho(j)}\Bigg)
    \end{aligned}
\end{equation*}
where $\rho(j)=j/2\sigma^2$ \cite{zhu2019poission,Mironov2019sampled}. By Lemma 3 of \cite{liang2020exploring}, we know that
\begin{equation*}
    \begin{aligned}
        D_\lambda 
        \Bigg( (1-\tau) &\mathcal{N} \Big(0 , \sigma^2  \Big) + \tau \mathcal{N}\Big(1, \sigma^2 \Big)  \Big\|\mathcal{N}\Big(0, \sigma^2 \Big)  \Bigg) \leq\\& 2\lambda \tau^2/\sigma^2 \leq \dfrac{\lambda \alpha_t L^2}{n^2},
    \end{aligned}
\end{equation*}
given $\sigma^2\geq 0.53$ and $\lambda-1 \leq \frac{2}{3}\sigma^2\log \big(1/\lambda\tau(1+\sigma^2)\big)$, where $\sigma^2 := \frac{2}{ \alpha_t \beta_t^2 L^2}$.

For step (2), we notice that the sequence $\{w_t\}_{t=1}^{T}$ can be seen as the outputs of a series of adaptive mechanisms
(see the iterative rule in Equation~\eqref{eq:sgld_update}). Here "adaptive" means the input of the current mechanism is
the output of the last mechanism. Thus we can further compose the bounds of interest via composition theories.
This way of composing privacy costs has been widely applied in prior works, such as moment accountants for differential privacy~\cite{dp_dl}.

In this paper, we borrow the composition theory from the prior work (Proposition 1 in \cite{renyi_dp}):
\begin{lemma} \label{lemma:compo}
 \textbf{(Composability)} Suppose 
 that a randomized algorithm $\mathcal{A}$ consists of a sequence
 of adaptive mechanisms $\mathcal{A}_1, \dots, \mathcal{A}_T$ where $\mathcal{A}_t:\prod_{j=1}^{t-1} w_j \times S \rightarrow w_t$. The \renyi divergence of the final outputs (on two adjacent datasets) can be bounded by:
 \begin{equation}
     D_{\lambda}(\mathcal{A}(S)\|\mathcal{A}(S')) \leq \sum_{t=1}^{T} D_{\lambda}(\mathcal{A}_t(S)\|\mathcal{A}_t(S')).
 \end{equation}
\end{lemma}
In our setting, the algorithm is set to be SGLD and the sub-mechanisms can compute the output
based on the updating rule at each iteration.
Combining Lemma~\ref{lemma:one_step_bound}  and \ref{lemma:compo}, we can obtain the final bound as the following theorem:
\begin{theorem}\label{theorem:sgld_bound}  
Suppose $w_T$ and $w'_T$ are the final outputs of SGLD on $S$ and $S'$, respectively. The training dataset  size is set to $n$. 
If the
gradient norm is bounded by $L$, then the \renyi divergence between
$p(w_T|S)$ and $p(w'_T|S')$ is bounded as:
\begin{equation}
    D_{\lambda}(p(w_T|S)\|p(w'_T|S')) \leq \dfrac{\lambda L^2 }{n^2}\sum_{t=1}^{T} \alpha_t.
\label{eq:sgld_bound}
\end{equation}
\end{theorem}
%\YY{Is it better to write the bound as $\tau^2\lambda/(1-\tau)\cdot*$ with $\tau=b/n$ is the subsampling rate? Then Thm 2 is at $O(\tau LC/\sqrt{1-\tau})$ and Thm 3 is at $O(\tau \sigma L/\sqrt{1-\tau} n^{-1/2})$ with additional $n^{-1/2}$.} 
Theorem~\ref{theorem:sgld_bound} is the main building block of our framework and we can further use it for deriving and interpreting previous results, which will be described in detail subsequently. From the view of privacy leakage,
Theorem~\ref{theorem:sgld_bound} demonstrates that under some mild conditions, the model learned using SGLD incurs a controllable privacy leakage, i.e., SGLD satisfies ($\lambda$, $\epsilon_n$)-\renyi differential privacy ($\epsilon_n$ is the right term of Equation~\eqref{eq:sgld_bound}). And the privacy cost $\epsilon_n$ is dominated by
the size of training dataset , i.e., $\epsilon_n=O(\dfrac{1}{n^2})$. In
Section~\ref{sec:exp}, we also empirically evaluate the information leakage against well-known
membership attack approaches.

\subsection{Stability-based Theory}
A promising way to study the generalization error is to derive the algorithm-dependent
generalization bounds via various stability notions~\cite{shwartz_stable,colt_sgld_gen,yuxiang_dp_stable}. The core idea is that the generalization error can be bounded in terms of the stability of the learning algorithm. A commonly-used notion is uniform stability, which can be connected to the expected generalization error of a randomized learning algorithm~\cite{shwartz_stable}. Mou et al.~\cite{colt_sgld_gen} propose to prove the uniform stability via bounding the squared Hellinger distance between $p(w_T|S)$ and $p(w'_T|S')$.
Here, the squared Hellinger distance between two density functions $p$ and $q$ are defined as: $D_H(p|q)\triangleq\dfrac{1}{2}\int(\sqrt{p}-\sqrt{q})^2dw$. 
We note that the boundness of the Hellinger distance can be obtained via the bound of
the \renyi divergence in terms of $D_H(p\|q)\leq D_\lambda(p\|q), \lambda\ge 1/2$.
Based on this observation, we can formally characterize the generalization error
as the following theorem:
\begin{theorem}
Consider the learning algorithm $\mathcal{A}$ to be SGLD with $T$ iterations and the batch size is set to  be $b$. And the sampling ratio $\tau$ is set to be $b/n$. Suppose that the loss function $l(z,w)$ is uniformly bounded by $C$, and the gradient
norm is bounded by $L$. The expected generalization error (Equation~\eqref{eq:gen_bound}) can be bounded as:
\begin{equation}
    \text{gen}(\mathcal{A},\mathcal{D}) \leq \dfrac{ \sqrt{2}LC}{n}(\sum_{t=1}^{T}\alpha_t)^{1/2}.
\end{equation}
\end{theorem}
\textbf{Proof Sketch:}
Denoting $\epsilon_n$ to be the right term in Equation~\eqref{eq:sgld_bound}. Then we obtain:
\begin{equation*}
    \begin{split}
        \text{gen}(\mathcal{A},\mathcal{D}) &\leq \sup_{S,S'}2C\sqrt{D_H(p(w_T|S)\|p(w'_T|S'))}\\ &\leq
    \sup_{S,S'}2C\sqrt{D_{1/2}(p(w_T|S)\|p(w'_T|S'))} \leq 2C \sqrt{\epsilon_n}.
    \end{split}
\end{equation*}
The proof details can be found in Appendix.
This generalization bound has a similar magnitude with the bound derived in the prior work \cite{colt_sgld_gen}. Different from \cite{colt_sgld_gen}, we take a novel view of privacy leakage analysis, which can largely simplify the derivation process and can be extended to a wide range of learning algorithms that satisfies
\renyi differential privacy, such as DP-SGD proposed in the prior work \cite{dp_dl}.
%Subsequent works propose to improve the generalization bound by introducing advanced stability notions, such as Bayes stability in \cite{nips_sgld_info}. 
\subsection{Information-based Theory}
The above discussion explores the connection between the information leakage and the stability and further obtains a specific bound for the expected generalization error.
We note that a more explicit way to quantify the information leakage is to measure the
mutual information between the learned weight and the training dataset. On the other hand,
we note that the main research line to study the generalization of a learning algorithm is to build the connection between mutual information and the expected generalization error.
Therefore, we can use mutual information to bridge the gap between privacy leakage and
the generalization bound of SGLD.
To this end, we introduce an important theoretical block used in this paper as the following lemma:
\begin{lemma} \label{lemma:gen_mi}
 (Proposition 1 in \cite{isit_sgld}) Suppose the loss function $l(z, w_t)$ is $\sigma$-subgaussian under the population distribution $\mathcal{D}$ for all $w\in \mathcal{W}$~\footnote{A random variable $X$ $\sigma$-subgaussian if $\log \mathbb{E}[e^{\lambda (X-\mathbb{E}X)}]\leq \lambda^2\sigma^2/2$  for all $\lambda\in\mathbb{R}$.}. $\mathcal{A}$ is the learning algorithm. Then we have:
 \begin{equation*}
      |\text{gen}(\mathcal{A},\mathcal{D})|\leq \dfrac{1}{n}\sum_{i=1}^{n}\sqrt{2\sigma^2 I(\mathcal{A}(S);z_i)}.
 \end{equation*}
\end{lemma}
The above lemma has established the formal connection between the mutual information and the generalization error and has been widely used in previous works that study the generalization of SGLD. In this paper, we further build the connection between the mutual information and the information leakage, which is reflected by the \renyi divergence in our framework.
Motivated by recent works~\cite{vi_mi_icml}, we propose using variational inference to derive the upper bound of $I(p(w_T|S);z_i)$. The derived bound can be further related to the
\renyi divergence between the posterior distribution over two adjacent datasets, i.e., 
$D_{\lambda}(p(w_T|S)\|p(w'_T|S'))$. This connection is formalized as:
\begin{lemma} \label{lemma:va_bound_mi}
 Suppose $S$ and $S'$ are adjacent and $\mathcal{A}$ is the learning algorithm, then:
 \begin{equation*}
     I(\mathcal{A}(S);z_i)\leq\mathbb{E}_{S}[D_\lambda(p(w_T|S)\|p(w'_T|S'))].
 \end{equation*}
\end{lemma}
The proof follows two steps: (1) deriving the variational bound of the mutual information using variational inference; (2) deriving the upper bound of the variational bound explicitly. The proof details are provided in Appendix. 
Combining Theorem~\ref{theorem:sgld_bound}, Lemma~\ref{lemma:gen_mi}, and Lemma~\ref{lemma:va_bound_mi}, we can build the theoretical connection
between the expected generalization error and the information leakage measured by the \renyi divergence as:
\begin{theorem}
Consider the learning algorithm $\mathcal{A}$ to be SGLD with $T$ iterations and the batch size is set to $b$. And the training dataset size is set to $n$. Suppose that the loss function $l(z,w)$ is $\sigma$-subgaussian under the population distribution $\mathcal{D}$ for all $w\in \mathcal{W}$. And the gradient
norm is upper bounded by $L$. The expected generalization error (Equation~\eqref{eq:gen_bound}) can be bounded as:
\begin{equation}
    |\text{gen}(\mathcal{A},\mathcal{D})| \leq  (\dfrac{\sqrt{2}\sigma L}{n})(\sum_{t=1}^{T}\alpha_t)^{1/2}.
\end{equation}
\end{theorem}
The derivation of the theorem paves a way to derive the generalization bound by the connection
between the mutual information and the information leakage (see details in Appendix). 

\if 0
\textbf{From the view of privacy analysis, we can obtain generalization bounds similar to previous works but with more simple and general proof. For example, we have obtained a generalization bound of $O(\dfrac{LC}{n})$ similar to \cite{colt_sgld_gen}. With the help of theoretical tools from the data privacy community \cite{renyi_dp}, we can have a simple proof to this bound while \cite{colt_sgld_gen} needs to carefully analyze the bound based on
stochastic differential equations (SDE) and transform the continuous setting into discrete one. More importantly, base on the technique used by \cite{colt_sgld_gen}, it is challenging to further improve the generalization bound. However, this can be easily done under our framework by introducing more advanced stability  notions or privacy notions. For example, we can replace the stability notion with Bayes stability used in \cite{Li2020On} to obtain a more tight bound.}
\fi

\subsection{Discussion on Our Results}

\paragraph{We provide a unified and succinct way to analyze the generalization ability of SGLD through information leakage analysis.}

The information leakage routine utilized in this paper unifies  two seemingly unrelated ways to obtain the generalization bound: the stability-based route and the information theoretic route. Meanwhile,  our approach largely simplifies the analysis:  For example, 
we have obtained a generalization bound of $O(\frac{LC}{n})$ similar to that of  \cite{colt_sgld_gen}, but with  a simple proof thanks to the  theoretical tools from the data privacy community \cite{renyi_dp}. 
However, \cite{colt_sgld_gen} need to  analyze the bound based on
stochastic differential equations (SDE) and transform the continuous setting into discrete one.

\paragraph{The generic privacy notion (\renyi differential privacy) adopted gives rise to refined bounds, and existing bounds could be further improved by using other generic notions, such as the Bayesian differential privacy.}

More importantly, base on the technique used by \cite{colt_sgld_gen}, it is challenging to further improve the generalization bound. However, this can be easily done under our framework by introducing more advanced stability  notions or privacy notions. For example, we can replace the stability notion with Bayes stability used in \cite{Li2020On} to obtain a more tight bound.

%The above discussions have pointed out two different ways to derive the generalization bound, i.e., using the stability-based and information theories. At a high level, the above two ways have derived the generalization bound in totally different manners. The stability-based methods bound the generalization error by observing individuals' impacts on the learned weight. While the information theory-based methods bound the generalization error by measuring the mutual information between the learned weight and the training dataset.  In this paper, we show that these two seemingly distinct ways can be unified by a view from the information leakage analysis. 

%Another notable aspect is the privacy notion used for analyzing information leakage. In this paper, we choose the \renyi differential privacy which is based on the \renyi divergence. This implies a potential way to improve the generalization bound by introducing a more advanced privacy notion. For example, one recent study \cite{bayes_dp} presents Bayesian differential privacy, which is to obtain the privacy bound in a  data-distribution-dependent manner. The Bayesian differential privacy can be seen as an improved version of \renyi differential privacy under some specific scenarios, which can provide more tight privacy bound.
%Thus we can replace the \renyi differential privacy with the Bayesian version to improve the generalization bounds obtained in this paper and previous works. We mark this as future work.

\subsection{Comparisons with Previous Works}
There are two most related direction to our works. One is to study generalization ability
of SGLD. Another is to study the connection between generalization and differential privacy in 
 general case. In this part, we highlight the differences between our work and these most related works.
 
 In contrast to research focusing on the generalization bounds of SGLD, we aim to unify this line of works through the information leakage analysis of SGLD. This unified view can help to
 simplify the derivation of generalization bounds of SGLD and provide new insights to improve these bounds.
 
 There are another research line focusing on 
 building the connection between generalization bounds and differential privacy in the general case instead of deriving concrete generalization bounds \cite{DworkFHPRR15,BassilyNSSSU16}.  For example, there are a series of work on adaptive data analysis \cite{DworkFHPRR15,hardt2014preventing,russo2016controlling,bassily2021algorithmic,jung2019new}, which study  false discovery and generalization error led by  different  settings of adaptivity.  While both of the works rely on techniques from differential privacy, in a specific case of SGLD, we need to carefully derive its information leakage
 in terms of \renyi differential privacy under some mild conditions, i.e., Theorem 1 and its derivation. Once we obtain its privacy leakage, we can employ previous works to build
 concrete generalization bound.

\section{Related Work}
\label{sec:related}
In this part, we briefly review some related works on the theoretical properties of SGLD.

\nosection{Convergence Property}
In early times, SGLD is proposed to enable Bayesian learning on large scale datasets and can be seen as an SG-MCMC method, which can approximate the process of sampling from the intractable posterior distribution (e.g., $p(w|S)$ in our paper).
One of the important properties of SGLD is that the parameters learned using SGLD will approach samples from the parameter
posterior distribution. Welling et al.~\cite{sgld_original} have shown some intuitive observations of this property.
Subsequent works have theoretically proved this proposition and have presented various convergence analysis, i.e., how fast the learned parameter can approach the posterior distribution. The key idea of these works is to treat SGLD as the discretization of some specific stochastic differential equations~(SDE, e.g., Fokker-Planck equation) and turn to study the discretization error~\cite{higher_order,sde_view_sgld}.

\nosection{Generalization Bounds of SGLD} 
More recently, the generalization bound of SGLD has emerged as an active topic. Most of the previous works study this problem from two theoretical aspects, the stability-based theory and information-theoretic quantities.
A representative work based on stability-based theory is conducted by Mou et al.~\cite{colt_sgld_gen}, in which they have derived that SGLD satisfies $\epsilon_n$ uniform stability with respect to the Hellinger loss. Then the expected generalization error is bounded by $\epsilon_n$ according to the stability-based theory \cite{shwartz_stable}. 
%A PAC-Bayesian bound is also obtained from the view of the discrete-time SGLD which treats the parameter updates as the solution of a specific PDE. 

Another research line studies the generalizability of SGLD from an information-theoretic view. 
The core idea is to bound the generalization error via some
information measures. For example, an early work \cite{XuR17} proposes to bound the expected generalization error in terms of the mutual information between the training dataset and the output hypothesis of the learning algorithm.
Bu et al.~\cite{isit_sgld} introduce the individual sample mutual information and touch it to the expected
generalization error. 
Besides these theoretical works, there is also empirical evidence showing that SGLD can improve the generalization ability of the model. For example, Gan et al. \cite{sgld_language_model} propose using SGLD to mitigate the overfitting of RNN models for language modeling tasks.

\nosection{Privacy Issues in SGLD}
There are also some related works that study the privacy issues in SGLD. Wang et al.~\cite{privacy_for_free_icml} present a variant of SGLD 
that satisfies differential privacy. Wu et al.~\cite{memberprivacy_sgld} further prove that SGLD can naturally prevent the well-known membership
attack  which is also conducted in our experiments. These works also point out some implicit connections between privacy leakage and generalizations. In our paper, we further formally build the connection
between previous theoretical results via the privacy leakage analysis.

%\nosection{Relation to Adaptive Data Analysis}

% the behavior of algorithms  in an adaptive interaction with the data.  

\section{Empirical Results}
\label{sec:exp}
In this section, we perform extensive experiments to empirically evaluate the information leakage of SGLD against membership attacks. In what follows, we first describe the detailed experimental settings, including descriptions of the datasets, models, and the attack method. Then we
demonstrate and analyze some numerical results. At last, we provide some discussions for further insights.
\begin{table*}
\centering
\caption{Overall results on three benchmarks. 
Attack denotes anti-attack ability measured by the accuracy of the membership predictor (\textbf{lower is better}). 
% \textbf{Attack denotes anti-attack ability measured by the AUC of the membership predictor (\textbf{higher is better}). }
Model denotes the model performance measured by
the accuracy of the test dataset (higher is better). Gap is the difference between training and test accuracy.}
\label{tab:results}
\begin{tabular}{cccccccccc}
\toprule
\multicolumn{1}{c}{\multirow{2}{*}{Method}} & \multicolumn{3}{c}{
% Fraud detection
German Credit
} & \multicolumn{3}{c}{UCI-adult} & \multicolumn{3}{c}{IDC} \\
\multicolumn{1}{c}{}                        & Attack & Model  & Gap  & Attack   & Model   & Gap  & Attack   & Model  & Gap  \\
\midrule
SGD                                        
%Plan B & 0.646        &0.734        &0.266    
& 0.646        &0.734     &0.266     
&  0.999       &0.849     &0.018
&0.646         &0.823     &0.170      \\
%PlanB &0.527          &0.747        &0.112
Dropout                                    
&0.527         &0.747       &0.112    
&0.891         &0.834       & 0.005 
&0.634         &0.818       &0.176      \\
%Plan B &0.539         &0.736        &0.075 
SGLD                                        
&0.539         &0.736     &0.075 
&0.609         &0.847     & 0.010
&0.620         &0.817     &0.156      \\
\bottomrule
\end{tabular}
\end{table*}
\subsection{Settings}
\nosection{Datasets}
The goal of our experiment is to empirically verify the connection between the generalization error and the privacy leakage of SGLD. Therefore we choose datasets from scenarios where data privacy is important, such as financial and census data analysis. Specifically, we select three datasets: 
% (1) The credit card fraud dataset for fraud detection~\cite{fraud_detection_data}. 
(1) The German Credit dataset for credit modeling.
(2) The UCI-adult dataset for 
income prediction~\cite{kohavi1996}. (3) The IDC dataset for pathological image classification \footnote{{\tt http://www.andrewjanowczyk.com/use-case-6\\-invasive-ductal-carcinoma-idc-segmentation/}}. 

% The credit card dataset consists of 284,807 transactions of credit cards. Each transaction has 28 features and is labeled as positive if the transaction is conducted by a fraud. We consider the binary classification task to identify ``fraud" transactions. We randomly split the whole dataset
% into training (57075 applications), hold-out/validation (114150 applications), and test~(57074 applications) sets. Here, the hold-out dataset can be used for either validation or building the ``shadow'' attack model~\cite{s_and_p_attack}.  
% Plan B 
The German Credit dataset consists of $1,000$ applications for credit cards. Each application is labeled with good or bad credit. We consider the classification task to identify ``good credit" applications. We randomly split the whole dataset into training ($400$ applications), hold-out/validation ($300$ applications), and test~($300$ applications) sets. Here, the hold-out dataset can be used for either validation or building the ``shadow'' attack model~\cite{s_and_p_attack}.  

We consider a second Adult Census Income dataset~\cite{kohavi1996} with 48,842 instances from UC Irvine repository \footnote{{\tt
http://archive.ics.uci.edu\\/ml/datasets/Census+Income
}} to examine the property attack issues. The dataset has 14 features such as country, age, work-class, education, etc. The goal for this binary classification task is to predict whether income exceeds $50K/yr$ based on the census information.  We split the whole dataset into training, validation, and test sets, each with \#22,792, \#9,769, \#16,281 instances respectively. The validation set is also used as the shadow dataset for building the attack model.

We also use a medical image dataset to observe the performance of convolutional neural networks (CNNs). 
Here, we consider the IDC dataset of pathological images for 
invasive ductal carcinoma (IDC) classification. This dataset contains $277\rm{,}524$ patches of $50\times50$ pixels ($198\rm{,}738$ IDC-negative and $78\rm{,}786$ IDC-positive). We split the whole dataset into training, validation~(hold-out), and test sets. To be specific, the training dataset consists of $10,788$ positive patches and $29,164$ negative patches. The test dataset consists of 
$11,595$ positive patches and $31,825$ negative patches. The remain patches are used as the hold-out dataset.

\nosection{Model Setup} 
% For comparison, we train the model using both SGLD and SGD variants. All these training strategies share the following hyper-parameters: the mini-batch is set to 32 and the epoch number is set to 30. The learning rate decreases by half every 5 epochs. For SGLD, the variance $\sigma^2$ of the prior is set to $1.0$. The initial learning rate is set to $0.001$. 
For the German Credit dataset, we train a three-layer fully-connected neural network for the classification task. For comparison, we train the model using both SGLD and SGD variants. All these training strategies share the following hyper-parameters: the mini-batch is set to be 32 and the epoch number is set to be 30. The learning rate decreases by half every 5 epochs. The initial learning rate is set to be $1\times10^{-3}$. 

For the UCI-adult dataset, we train a fully-connected neural network with a structure of [32, 16].  The mini-batch size and the number of epochs are set as 64 and 100 respectively.
SGD is adopted with initial learning as 0.1 and halves every 20 epochs, while 
for SGLD, the variance $\sigma^2$ of the prior is set to be 1.0. 

For images in the IDC dataset, we train a ResNet-18 model for the classification task (i.e, idc and non-idc). The mini-batch size
is set to be 128 and the epoch number is set to be 100. Data augmentation is not used. The learning rate decreases by half every 20 epochs.
For SGLD, the variance $\sigma^2$  of the prior is set to be 1.0. The initial learning rate is set to be $1\times10^{-4}$.

\nosection{Attack Setup}
To quantitatively evaluate the information leakage of different learning algorithms, we perform membership attacks on learned models. Given a learned model, the goal of the membership attack is to determine whether a sample is used for training the model.
Formally, we can treat this task as a binary classification problem, in which we need to train a membership predictor to predict the membership information of a sample (i.e., belonging to the training dataset or not). 
In this paper, for %fraud detection 
German Credit and UCI-adult datasets, we employ a widely used technique named \emph{shadow model training} for building the membership predictor.
We refer readers to the prior work~\cite{s_and_p_attack} for more details. For the IDC dataset, we use the threshold attack following the prior work~\cite{memberprivacy_sgld}. We use the accuracy of the membership attack to measure the privacy leakage of different training strategies (see the details of the training strategies in the next section). 
\subsection{Result}

\begin{table}
\centering
\caption{The results of property attack against transfer learning on the UCI dataset. The target property is set to "race".}
\label{tab:property_attack}
\begin{tabular}{ccc}
\toprule
Strategy &Model &Attack \\ \hline
SGD &0.846   &0.777 \\
SGLD &0.850  &0.686\\\bottomrule
\end{tabular}
\end{table}
Table~\ref{tab:results} shows that the information leakage of three different training strategies. To be specific, Dropout denotes adding Dropout layers into the network architecture. For the fully-connected network used for German Credit and UCI-adult, we insert Dropout after all hidden layers and set the drop ratio to 0.5. For ResNet-18, we insert Dropout between convolutional layers and set the drop ratio to $0.3$ following the work~\cite{p3sgd}.

As shown in Table~\ref{tab:results}, SGLD has demonstrated strong anti-attack ability and generalization ability in terms of the attack accuracy and the model accuracy, respectively. 
For the UCI-adult task, in comparison to the conventional SGD algorithm, the model trained using SGLD achieves better generalization ability in terms of the gap between the training and test model accuracy while reducing the attack accuracy score from 0.999 to 0.609. 
For the IDC task, SGLD has also demonstrated its ability in reducing the information leakage (the attack accuracy score decreased from 0.646 to 0.620) while achieving better generalization ability than SGD (better model accuracy and smaller training-test accuracy gap). Previous works have shown Dropout can be seen as an effective method to improve the generalization ability of the learned model. In our experiment, we also validate its ability to reduce information leakage in the case of a fully-connected network.  

\nosection{Transfer learning} We note that there are naturally two data parties in transfer learning, i.e., the source and target domains. And a commonly-used approach of transfer learning is to build the training protocol based on feature alignment loss~\cite{long2015learning}. This approach involves the hidden feature communications between two data parties, which incurs information leakage risks. Here we show that SGLD can reduce such an information leakage while achieving a comparable model accuracy. Specifically, we consider the UCI dataset in the context of transfer learning, where we use data instances with the country of “U.S.” as a source domain and “non-U.S.” as the target. An MLP is used as the base model. To evaluate the information leakage, we perform the property attack that uses the "race" as the target property (see more details of the property attack in \cite{melis2019exploiting} ). The results are shown in Table~\ref{tab:property_attack}, which demonstrates SGLD can reduce the information leakage incurred by the hidden feature communications in terms of the attack accuracy.

% \begin{tabular}{llllllllll}
% \toprule
% \multicolumn{1}{c}{\multirow{2}{*}{Method}} & \multicolumn{3}{c}{Fraud detection} & \multicolumn{3}{c}{UCI-adult} & \multicolumn{3}{c}{IDC} \\
% \multicolumn{1}{c}{}                        & Attack  & Model  & Gap  & Attack   & Model   & Gap  & Attack  & Model  & Gap  \\
% \midrule
% SGD                                        
% &         &        &    
% &          &  0.902/0.849     &0.014/0.008
% &         &0.823        &0.170      \\
% Dropout                                    
% &         &        &      
% &          & 0.877/0.827      &   0.007/-0.001  
% &         &0.818       &0.176      \\
% SGLD                                        
% &         &        &    
% &         &0.898/0.844     & 0.013/0.010
% &         &0.817        &0.156      \\
% SGLD+Clip                                  
% &         &        &      
% &          & 0.862/0.829        &  0.002/0.000
% &         &        &      \\
% \bottomrule
% \end{tabular}
% \end{table}
% \subsection{Discussion}

\section{Conclusion}
In this paper, we study the generalization bound of Stochastic Gradient Langevin Dynamics from the perspective of privacy leakage analysis.
Based on this new perspective, we build the theoretical framework to derive the generalization bound of SGLD. Our framework can provide a unified interpretation of previous theoretical results that are based on information/stability-based theories. Moreover, our research can shed a light on improving the generalization bound by using a more advanced privacy notion. 
\bibliographystyle{aaai}
\bibliography{refs}
\end{document}